# Human Gait Analysis using Gait Energy Image

Sagor Chandro Bakchy[1], Md. Rabiul Islam[2], M. Rasel Mahmud[3] and Faisal Imran[4]
Department of Computer Science and Engineering,
Rajshahi University of Engineering and Technology[2] and Varendra University[1,3,4], Rajshahi-6204, Bangladesh
sagorchandro.10@gmail.com, rabiul_cse@yahoo.com, m.raselmahmud1@gmail.com, imran@vu.edu.bd

**Abstract--***Gait recognition is one of the most recent emerging techniques of human biometric which can be used for security based purposes having unobtrusive learning method. In comparison with other bio-metrics gait analysis has some special security features. Most of the biometric technique uses sequential template based component analysis for recognition. Comparing with those methods, we proposed a developed technique for gait identification using the feature Gait Energy Image (GEI). GEI representation of gait contains all information of each image in one gait cycle and requires less storage and low processing speed. As only one image is enough to store the necessary information in GEI feature recognition process is very easier than any other feature for gait recognition. Gait recognition has some limitations in recognition process like viewing angle variation, walking speed, clothes, carrying load etc. Our proposed method in the paper compares the recognition performance with template based feature extraction which needs to process each frame in the cycle. We use GEI which gives relatively all information about all the frames in the cycle and results in better performance than other feature of gait analysis.*

## I. INTRODUCTION

Gait is a human biometry that indicates human unique walking style, body movements, and the activity of muscles at the time of walking or running. Human walking is a periodic incident. Starting from stepping the right leg then putting the left leg and again putting the right leg on the ground is considered as one gait cycle or gait period. Human has 32 sets of gait feature data like stribe, torso, hand, angle of joints, distance of foots and foot length etc. Gait analysis includes image processing like other bio-metric technique. In recognizing people using gait analysis first video is captured from distance with a static camera. Gait recognition approach is mainly divided into two major categories as model based approach and motion based approach [1]. Model based approach focuses on area related information likely stride, torso while motion based approach works with silhouette. Silhouette is the vivid portion of foreground in front of a dim background and is extracted from the video [2]. GEI is then found from the silhouettes. Silhouettes are extracted from video frame captured by camera using background subtractions. Gait analysis implementation has three basic types. These are i) Machine Vision (MV), ii) Floor Sensor (FS), iii) Wearable Sensor (WS). Machine vision based technique is the most common way that captures video of the surveillance area [3]. Floor sensor detect motion of human when he/she walks over the sensor mat on the ground. People wear a sensor in the part of leg and respective data are recorded. In GEI based recognition each extracted silhouette contains average information about several frames in the nearby frames [4]. Gait recognition performance is affected by some external facts like wearing trouser, carrying load, viewing angle and some internal fact like temporary or permanent injury, drunkenness and pregnancy.

## II. RELATED WORKS

Concept of gait analysis is very old. Aristotle (384–322 BCE) is the person who commented about the manner in which human walk. Willhelm(1804–1891) and Eduard (1806–1871) Weber, working in Leipzig who made the next major contribution based on very simple measurements. Verne Inman (1905–1980) and Howard Eberhart (1906–1993) made major advances in America shortly after the Second World War. In recent years the first and initially most popular mechanism for measuring gait was computer vision. One vision approach proved to be practical for tracking people in large, open spaces if cameras can be placed high above the area. Another project uses Markov models to sequence the various postures of a person while walking.Computer vision has also been used for people counting systems [5]. The vision approaches are generally limited by the requirement for a specific camera vantage point or other physical attribute. Thus, in cases where the vision requirements cannot be met, other approaches are more practical. One such method utilizes continuous wave radar [6]. The strength of the wave radar approach is to determining the presence or absence of humans. The work in this paper focuses a step further on how to identify specific people after the detection of a person has occurred. Many related projects use accelerometers to identify a person based on gait. Accelerometers are attached to the person either at the leg or the small of the back. All these approaches, use accelerometer data, not gyroscope data for recognition purpose using gait analysis [7]. Also Gafurav's method uses only use accelerometer data for gait recognition rather than gyroscope data [8].

## III. PROPOSED METHODOLOGY

Our proposed methodology for gait recognition uses the CASIA-B multiview dataset. The dataset contains preprocessed silhouette for training into the machine. Our system has training and tgesting period. In training period, video of the surveillance area was captured by using a static camera. Then frames were extracted from the video. All possible frames were selected in one complete gait cycle and GEI was extracted from those selected frames. The final GEI was learned into the machine as reference database.

In testing period, video frames were input from CASIA-B multiview database as experimental setup was simulation based. The GEI of gait cycle was extracted and similarity with stored GEI in the database was measured. Finally, the recognition result was obtained. The proposed methodology can identify and verify person in the surveillance area. When GEI of testing period was matched with GEI in database, the person was identified. Otherwise the person is verified as unauthorized entry. The main concept of our methodology is shown in figure 1.

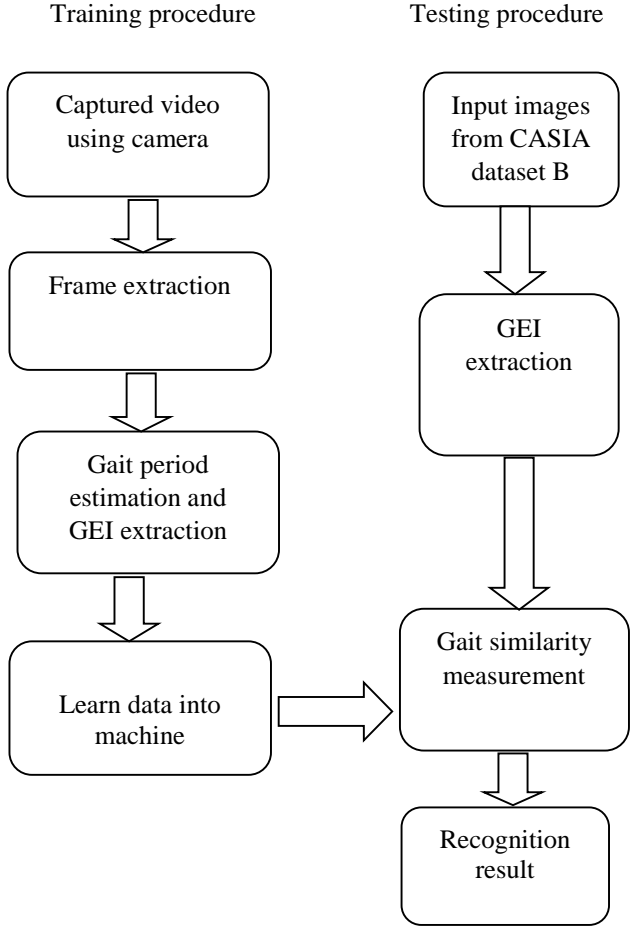

Figure 1: The proposed methodology.

## IV. IMPLMENTATION

GAIT ENERGY IMAGE

Human walking style is a periodic nature. Each cycle contain several frames. Gait Energy Image (GEI) is mainly average of all silhouete in one gait cycle [9]. GEI represent most possible information of the image sequences. GEI is less affected by the noise as only one image to be processed.

$B_t(x, y)$ at time t in a sequence, the gait energy image (GEI) is defined as follows:

$$G(x,y) = \frac{1}{N}\sum_{t=1}^{N} B_t(x,y) \quad (1)$$

Where N is the number of frames in one complete gait cycle(s) of a walking person, t is the frame number in the sequence (moment of time), and x and y are values in the 2D image coordinate. If we consider a noisy silhouette image, $B_t(x, y)$ that is formed by the addition of noise $\eta_t(x,y)$ to an original silhouette image $f_t(x,y)$ that is $B_t(x, y) = f_t(x,y) + \eta_t(x,y)$. Under these constraints, we further assume that $\eta_t(x,y)$ satisfies the distribution that is described by the following equation:

$$\eta_t(x,y) = \begin{cases} \eta_{1t}(x,y): P\{\eta_t(x,y) = -1\} = p, \\ P\{\eta_t(x,y) = 0\} = 1 - p, if\ f_t(x,y) = 1 \\ \eta_{2t}(x,y): P\{\eta_t(x,y) = 1\} = p, \\ P\{\eta_t(x,y) = 0\} = 1 - p, if\ f_t(x,y) = 0 \end{cases} \quad (2)$$

We have

$$E\{\eta_t(x,y)\} = \begin{cases} -p, & if\ f_t(x,y) = 1 \\ p, & if\ f_t(x,y) = 0 \end{cases} \quad (3)$$

Example of GEI extraction from a sequence of image is given in figure 2. Here four images from one gait cycle was taken and the GEI found at last which contains information of all portion of one complete gait cycle.

FEATURE EXTRACTION

GEI was used as feature in the experiment. As GEI contains all information of complete gait cycle, one frame was enough for recognition process. To ensure perfect GEI extraction the only body potionrt of a human being (Region of Interest, ROI) was extracted from the background. Training machine with such one frame was comparatively easy than other methods. Besides, less storage memory and processing time were required. Input images were selected and then checked with the trained database.

## V. EXPERIMENTAL RESULT

Gait recognition was done using the sample data. But real life data may cause difficulty to the recognition process. As the outcome we have found, the rate of recognition is quite better than the other existing method for gait analysis.Required time for recognition decreases in a large scale. The time comparision for data matching in template based and GEI based method is shown in the following table:

Table 1: Time comparison of Template based and GEI Based recognition.

| Method | No. of person | Image in each sequence | Time needed (s) | Total image processed |
|---|---|---|---|---|
| Template | 32 | 352 | 2.84 | 352 |
| GEI | 32 | 352 | 1.61 | 32 |

Time needed in template based recognition, t(m)=2.84 s
Time needed in GEI based recognition, t(g)=1.61 s

Different of time, t(d) = {t(m)-t(g)} s
= (2.84-1.61) s
= 1.23 s
Time reduced = (1.23/2.84) *100%
= 43.31%

So, time efficiency is 43.31% in GEI based method than template based method.

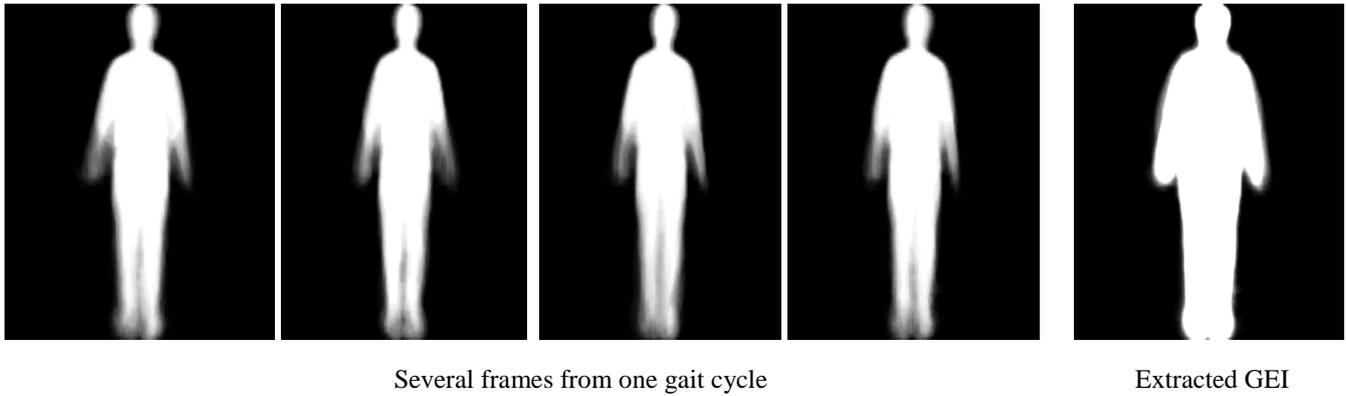

Several frames from one gait cycle      Extracted GEI

Figure 2: GEI Extraction of the frames in one complete gait cycle

Table-2: Gait recognition rate using GEI.

| Trained Data | Tested Data(a) | Recognized Data (b) | Rate c=b/a*100 |
|---|---|---|---|
| 10 | 09 | 05 | 55.55 |
| 32 | 23 | 13 | 56.52 |
| 45 | 35 | 21 | 60.02 |

We had trained machine with different number of people at different time. Each time result was different depending on the number of total image in the database. We found nearly 60% accurecy rate in our experiment. Average gait recognition result was found as listed in table 2.

The comparison of performance with other method is shown in figure 3, which reveals that our proposed methodology has improved performance than some other established methods.

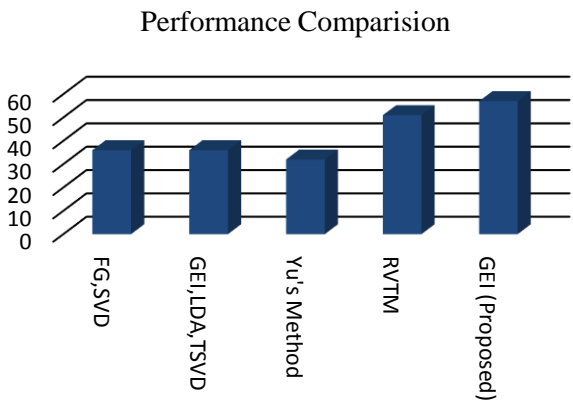

Figure 3: Gait Recognition Rate comparison.

The FG and Singular Value Decomposition (SVD), Gait Energy Image (GEI) -TSVD both method has nearly 36% accuracy rate. Yu's method has recognition rate of 32% and Robust View Transformation Method (RVTM) has 51% accuracy. Ours proposed methodology has nearly 57% of recognition rate [10].

## VI. CONCLUSION

We have just tested with the sample data for recognition process. Due to use sample data, recognition rate was very high. In future, the system will be developed by increasing the number of testing data to attain an improved recognition result. Real time video capturing and feature extraction for real data will be designed. In future this work will be done to ensure a sophisticated real type system implementation. Also recognition of multiple in the same frame will be considered in future.

.